\documentclass[runningheads]{llncs}
\usepackage{graphicx}
\usepackage{subfigure}
\usepackage{subcaption}
\usepackage{multirow}
\usepackage{booktabs} 
\usepackage{array}

\usepackage{amsmath}

\DeclareMathOperator*{\argmin}{arg\,min}

\newcolumntype{C}[1]{>{\centering\arraybackslash}m{#1}}
\begin{document}
\title{Graph Edits for Counterfactual Explanations: A comparative study}
\author{Angeliki Dimitriou\orcidID{0009-0001-5817-3794}* \and
Nikolaos Chaidos\orcidID{0009-0006-0347-2785}* \and
Maria Lymperaiou\orcidID{0000-0001-9442-4186} \and
Giorgos Stamou\orcidID{0000-0003-1210-9874}}
\authorrunning{A. Dimitriou et al.}
\institute{National Technical University of Athens
\email{gstam@cs.ntua.gr}\\
\email{\{nchaidos,angelikidim,marialymp\}@ails.ece.ntua.gr}\\
* These authors contributed equally}
\maketitle              % typeset the header of the contribution

\begin{abstract}
Counterfactuals have been established as a popular explainability technique which leverages a set of minimal edits to alter the prediction of a classifier. When considering conceptual counterfactuals on images, the edits requested should correspond to salient concepts present in the input data. At the same time, conceptual distances are defined by knowledge graphs, ensuring the optimality of conceptual edits.
In this work, we extend previous endeavors on \textit{graph edits as counterfactual explanations} by conducting a comparative study which encompasses both supervised and unsupervised Graph Neural Network (GNN) approaches. To this end, we pose the following significant research question: should we represent input data as graphs, which is the optimal GNN approach in terms of performance and time efficiency to generate minimal and meaningful counterfactual explanations for black-box image classifiers?
%: should we represent input data as graphs, which is the shortest graph edit path that results in an alternative classification label as provided by a black-box classifier?  

\end{abstract}

\section{Introduction}
In the era of large and complex neural models, ensuring trust between them and humans becomes a critical issue. Explainability literature suggests a variety of methods to probe neural models behaviors, either requiring access to their inner workings \cite{goyal2019counterfactual,vandenhende2022making} or not \cite{dervakos2023choose,dimitriou2024structure,cece}. There is also ample discussion regarding the nature of the features involved in a human-understandable explanation; for example, low-level features e.g. pixel-related characteristics (brightness, contrast) may be unable to provide a meaningful explanation to the end user, despite being informative for a neural model \cite{rudin2019stop}. This observation is applicable in the counterfactual explanation scenario, suggesting that semantics are fundamentally essential to meaningful counterfactual explanations \cite{browne2020semantics,MILLER20191}. In the meanwhile, both high-level semantics as well as low-level features can be expressed in the same mathematical format (e.g. numerical vectors), allowing the transition to semantically rich explanation systems \cite{dervakos2023choose}. In the interest of explaining visual classification systems, and with respect to recent related work \cite{cocox,dervakos2023choose,dimitriou2024structure,cece,goyal2019counterfactual,vandenhende2022making}, we base our current study on \textit{conceptual counterfactual explanations}. 

\begin{figure}
    \centering
    \includegraphics[width=0.87\linewidth]{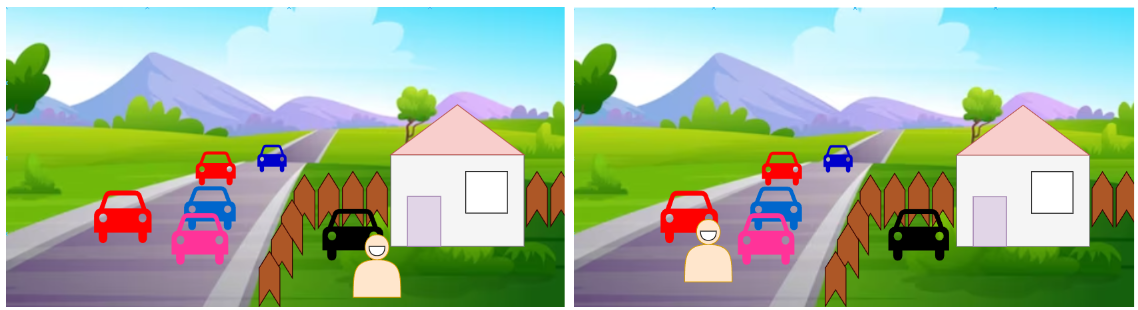}
    % \caption{A safe situation of a person in front of a parked car (left) compared to a non-safe situation of a person in front of moving cars on the highway (right).}
    
    % \caption{Left - person in front of parked car (safe). Right - person in front of moving cars on the highway (unsafe). The relationship between the person, the car and the highway is critical for the transition to the counterfactual class.}

    \caption{Left - person in front of parked car (safe). Right - person in front of moving cars on the highway (unsafe). The relationship between the person, the car and the highway is critical for the transition to the counterfactual class.}
    \label{fig:example-graph}
    \vspace*{-0.75\baselineskip}
\end{figure}

With regard to the model accessibility an explainability system can have, most related counterfactual explanation methods belong to the white-box category \cite{cocox,goyal2019counterfactual,vandenhende2022making}, offering insights tailored to the model under investigation. However, in the emergence of powerful, yet proprietary models, such as ChatGPT \cite{chatgpt} and GPT-4 \cite{gpt4} which are currently accessible only through APIs, black-box explanations resurface as more viable and universal solutions, able to approach a model by scrutinizing their outputs. Under these constraints, we stay within the \textit{black-box} setting of conceptual counterfactuals \cite{dervakos2023choose,dimitriou2024structure,cece}.

\begin{figure*}[t!]
    \centering
    \includegraphics[width=\textwidth]{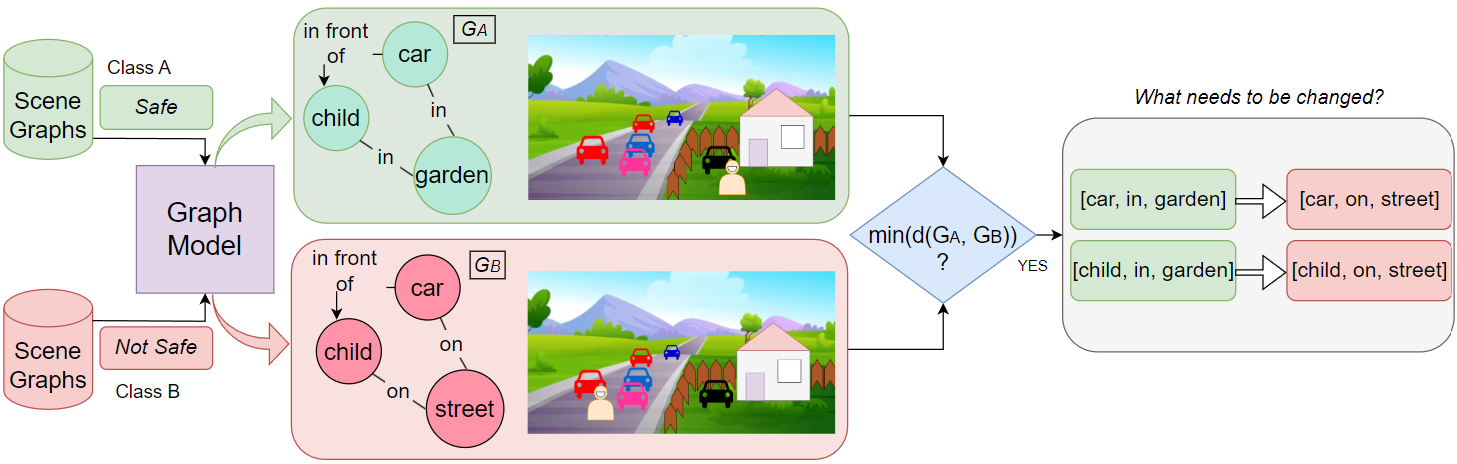}
    \caption{Outline of our evaluation framework: Given scene graphs of classes $A\neq B$, a graph model embeds them in a low-dimensional space, allowing the retrieval of the closest graphs $G_A$, $G_B$ from which we extract counterfactual edits.}
    \label{fig:outline}
    \vspace*{-0.75\baselineskip}
\end{figure*}

We also acknowledge the importance of intra-concept roles, as associations between the same concepts via roles may result in different interpretations of situations (Figure \ref{fig:example-graph}). Thus, we advocate image representation via scene graphs. To this end, we extend the framework of \cite{dimitriou2024structure} to accommodate a larger spectrum of Graph Machine Learning algorithms to serve conceptual counterfactual explanations on scene images. We retain the brute-force calculation of Graph Edit Distance (GED)\cite{ged} as the ground truth counterfactual measure, under which the closest scene graph in terms of GED  belonging to a different class is considered as the ``gold'' counterfactual graph. In order to overcome the computational complexity of GED calculation, we employ Graph Machine Learning algorithms, such as Graph Kernels, Graph Autoencoders (GAEs) and supervised Graph Neural Networks (GNNs), which are lightweight and significantly faster than the NP-hard deterministic GED calculation \cite{ged_np_hard}, even when optimizations are employed \cite{jonker1987shortest}.
Graph edits serving as counterfactual explanations are instructed via concept distances in existing hierarchies, in our case being WordNet \cite{miller1995wordnet}. 
Merits of such explanations include but are not limited to actionability, a crucial quality of counterfactuals in several domains \cite{menis2024beyond}.
The outline of our approach is presented in Figure \ref{fig:outline}. Overall, we explore the capabilities of Graph Machine Learning algorithms, contributing to the following:

\begin{itemize}
    %\item We calculate \textit{graph-based conceptual counterfactual explanations} overcoming significant limitations of prior work on conceptual counterfactuals.
    \item We prove quantitatively and qualitatively that both unsupervised (GAEs) and supervised  (GNNs) methods can provide meaningful and accurate conceptual counterfactuals, viewing models under explanation as black-boxes.
    \item We demonstrate the trade-offs between unsupervised  vs supervised approaches towards counterfactual explanations, and provide insights on the advantages of each method.
\end{itemize}

\section{Related work}
\textit{\textbf{Counterfactual Explanations}}
have been established as an effective explanation method \cite{wachter2018counterfactual} should certain requirements, such as actionability and feasibility be met \cite{face}. These principles are transferred in the field of counterfactual visual explanations, which aim to explain the behaviors of image classifiers. Most of these approaches leverage pixel-level edits, denoting regions that should be minimally changed \cite{goyal2019counterfactual,hendricks2018grounding,vandenhende2022making} and often proceed to actually generate the targeted region by employing generative models \cite{augustin2022diffusion,chang2019explaining,Farid2023LatentDC,Zhao2020GeneratingNC}.

Since semantics are inherently tied to human-interpretable counterfactuals \cite{browne2020semantics}, a diverging line of work suggests moving towards conceptual rather than pixel-level counterfactuals. The notion of explainable concepts  was presented in \cite{cocox}, where fault-lines are utilized to drive concept edits. \cite{abid2022meaningfully} produce conceptual counterfactuals to meaningfully explain model misclassifications. However, these approaches require some degree of access to the internals of the classifier.
Minimal concept edits in a black-box fashion were proposed in \cite{cece}, paving the way for knowledge-based model-agnostic visual counterfactuals. Some limitations were resolved in subsequent works, suggesting an indirect \cite{dervakos2023choose} or direct \cite{dimitriou2024structure} incorporation of roles into concepts, while underlining the importance of correctly selecting data and representations to obtain proper explanations. The same idea is modified to explain hallucinations of image generation models rather than classifiers \cite{lymperaiou2023counterfactual}, highlighting the significance of universal black-box explainers.

\section{Method}
%\subsection{Background} 

The focus of our current study centers on determining the minimal semantic alteration necessary for an image $i$ to transition from being classified as A to B. Each image $i$ in our dataset $I$ is represented by a scene graph $G_i$, encapsulating semantic information, therefore forming an explanation dataset \{$i, G_i$\}, as previously outlined in \cite{cece}. Utilizing a reference image $i_r$ and candidate scene graphs $G_{i_c}$, the GED is computed using an optimization approach to identify the most efficient path for semantic transformation between $G_{i_r}$ and $G_{i_c}$, guided by predefined graph edit operations (Replacement, Deletion, Insertion). Semantic distances, drawn from the well-crafted WordNet \cite{miller1995wordnet} hierarchy, guide the choice of edits, ensuring meaningful transitions while penalizing irrelevant replacements. To accelerate GED calculations, we employ optimization techniques similar to \cite{dimitriou2024structure}, utilizing the Volgenant-Jonker (VJ) algorithm \cite{jonker1987shortest}. Finally, to obtain the counterfactual candidate $G_{i_{cc}}$ of the original scene graph $G_{i_{r}}$, we repeat the optimized GED calculation for each candidate $G_{i_{c}}$, where $c \in C=I \setminus \{i_r\}$, ensuring that the selected closest $G_{i_{cc}}$ belongs to class $B\neq A$, while $G_{i_r}$ belongs to class $A$, formulated as:

\begin{equation}
   i_{cc}=\argmin_{i_{c} \in C}\left(GED\left(G_{i_r}, G_{i_c}\right)\right), \;\;\;\;\;  G_{i_r} \in A,\; G_{i_c} \in B\neq A
\label{eq:counterfactual-summary}
\end{equation}

% The core of the current work revolves around the fundamental question of "What is the minimal \textit{semantic} change that has to occur in order for an image $i$ to be classified as B instead of A?". Each image $i$ in our dataset $I$ is represented by a scene graph $G_i$, comprising nodes and edges that contain \textit{semantics s},
% ultimately constituting an \textit{explanation dataset} \{$i, G_i$\}, $i \in I$ as in \cite{cece}.

% Considering a reference image $i_r$ with corresponding scene graph $G_{i_r}$, and candidates $G_{i_c}, c \in C= I \setminus \{i_r\}$, GED between $G_{i_r}$ and a random candidate is:
% \begin{equation}
%     \textit{GED($G_{i_r}$, $G_{i_c}$)}=\min_{(e_1,..., e_n)\in P(G_{i_r}, G_{i_c})}\sum_{i=1}^{n}c(e_i)
% \label{eq:eq1}
% \end{equation}
% where c(e) denotes the cost of a graph edit operation $e$ among n possible valid edits, and $P(G_{i_r}, G_{i_c})$ is the set of graph paths needed to transform $G_{i_r}$ to $G_{i_c}$. 

\subsection{The importance of Graph Machine Learning}
Considering a dataset with $N$ graphs, calculating the counterfactual $G_{i_{cc}}$ for every $G_{i_r}$ requires $O(N^2)$ GED operations, resulting in a prohibitive calculation time, despite using optimizations. In order to accelerate this process, we leverage Graph Machine Learning algorithms to circumvent the exhaustive calculation of $N^2$ GEDs, accepting an unavoidable performance sacrifice. 

\textit{\textbf{Graph Kernels}} is our first step towards testing graph matching \cite{kernels}. Kernels in Machine Learning are responsible for mapping the given data in high-dimensional spaces, revealing complex patterns that cannot be observed in the original dimensionality of the input data, therefore allowing the application of linear algorithms to solve non-linear problems. Graph kernels quantify the degree of similarity between subgraphs in polynomial time by employing kernel functions, thus contributing to the overall structural similarity between two given graphs. 
In our experiments, we will compare different graph kernel approaches.

\textit{\textbf{Graph Neural Networks (GNNs)}}
%Moving on to neural implementations, there are both supervised and unsupervised techniques able to provide efficient GED approximations. 
in the supervised setting comprise architectures such as Graph Convolution Network (GCN) \cite{kipf2016semi}, Graph Attention Network (GAT) \cite{velivckovic2017graph}, and Graph Isomorphism Network (GIN) \cite{xu2018powerful}. They are incorporated in a Siamese Network to learn similarity on input instances \cite{li2019graph},
demonstrating state-of-the-art results even when trained on a smaller subset of $N^2/j$ labeled graph pairs ($N$ is the number of graphs in the complete dataset, $j$ a positive integer). During training, these GNN variants learn the relationships between counterfactual graph pairs utilizing ground truth samples calculated from Eq. \ref{eq:counterfactual-summary}. Embedding representations on new graphs are extracted using the trained GNNs, translating graph similarity cues to distances in the embedding space, therefore allowing retrieval of counterfactual graphs $G_{i_{cc}}$ for each $G_{i_r}$.

\textit{\textbf{Graph Autoencoders (GAEs)}} \cite{kipf2016variational} arise as faster and more versatile solutions by eliminating the need for labeled samples, even though their metric performance often ranks lower than GNNs. GAEs implement an encoder-decoder structure, with the encoder being tasked to produce embedding representations; these representations are useful for retrieval tasks, since similarity in the embedding space corresponds to actual graph similarity. Thus, the need of GED calculations within the training process is completely avoided, dropping the computational complexity to $O(N)$ in comparison to the $O(N^2)$ operations required in the supervised case ($N$ is the number of graphs). The encoder and decoder modules are constructed using any of the aforementioned GNN variants (same GNN modules for the encoder and the decoder when applicable). 

In all cases, we retain the NP-hard GED scores for ground truth comparison to evaluate the performance of different models on the ground-truth rank.

\section{Experiments}

\begin{table}
\vspace*{-0.95\baselineskip}
\small
\centering
\caption{Counterfactual retrieval results with Supervised and Unsupervised GNN architectures. \underline{Underlined} cells indicate best results per graph model (kernel, supervised, unsupervised). \textbf{Bold} denotes best results overall per dataset split.}
\label{tab:combined-results}
\vskip 0.1in
\begin{tabular}{c|ccc|ccc|ccc}
\hline
Model & \multicolumn{3}{c}{NDCG (binary)} & \multicolumn{3}{c}{Precision (P) (binary)} & \multicolumn{3}{c}{Precision (P)} \\
\cline{2-10}
& @4 & @2 & @1 & @4 & @2 & @1 & @4 & @2 & @1 \\
\hline
\multicolumn{10}{c}{VG-DENSE} \\
\hline
WL kernel & 0.287 & 0.198 & 0.138 & 0.186 & 0.128 & 0.076 & \underline{0.164} & 0.125 & 0.076 \\
SP kernel & \underline{0.334} & \underline{0.251} & \underline{0.194} & \underline{0.232} & \underline{0.174} & \underline{0.122} & 0.154 & \underline{0.145} & \underline{0.122} \\
RW kernel & 0.0 & 0.0 & 0.0 & 0.0 & 0.0 & 0.0 & 0.001 & 0.0 & 0.0 \\
NH kernel & 0.282 & 0.192 & 0.131 & 0.124 & 0.086 & 0.05 & 0.088 & 0.073 & 0.05 \\
GS kernel & 0.259 & 0.166 & 0.103 & 0.018 & 0.008 & 0.008 & 0.022 & 0.007 & 0.008 \\
\hline
Superv. GAT & 0.342 & 0.26 & 0.204 & 0.422 & 0.288 & 0.164 & 0.312 & 0.25 & 0.164 \\
Superv. GIN & 0.339 & 0.256 & 0.2 & 0.344 & 0.242 & 0.158 & 0.252 & 0.192 & 0.158 \\
Superv. GCN & \textbf{0.407} & \textbf{0.333} & \textbf{0.283} & \textbf{0.492} & \textbf{0.37} & \textbf{0.248} & \textbf{0.358} & \textbf{0.292} & \textbf{0.248} \\
\hline
GAE GAT & 0.281 & 0.191 & 0.13 & 0.108 & 0.078 & 0.048 & 0.075 & 0.075 & 0.048 \\
GAE GIN & 0.281 & 0.191 & 0.13 & 0.132 & 0.098 & 0.052 & 0.082 & 0.078 & 0.052 \\
GAE GCN & 0.287 & 0.197 & 0.137 & 0.102 & 0.078 & 0.052 & 0.071 & 0.06 & 0.052 \\
VGAE GAT & 0.298 & 0.21 & 0.15 & 0.196 & 0.138 & 0.082 & 0.139 & 0.12 & 0.082 \\
VGAE GIN & 0.304 & 0.216 & 0.157 & 0.202 & 0.146 & 0.09 & \underline{0.141} & 0.119 & 0.09 \\
VGAE GCN & 0.287 & 0.198 & 0.137 & 0.174 & 0.118 & 0.066 & 0.122 & 0.101 & 0.066 \\
GFA GAT & 0.302 & 0.214 & 0.155 & 0.182 & 0.14 & 0.086 & 0.122 & 0.109 & 0.086 \\
GFA GIN & 0.309 & 0.223 & 0.164 & 0.21 & 0.148 & 0.096 & 0.14 & 0.122 & 0.096 \\
GFA GCN & 0.163 & 0.165 & 0.158 & \underline{0.289} & \underline{0.2} & \underline{0.139} & 0.12 & 0.104 & 0.068 \\
ARVGA GAT & 0.3 & 0.213 & 0.153 & 0.188 & 0.14 & 0.084 & 0.132 & 0.114 & 0.084 \\
ARVGA GIN & \underline{0.317} & \underline{0.232} & \underline{0.174} & 0.218 & 0.164 & 0.106 & 0.135 & \underline{0.124} & \underline{0.106} \\
ARVGA GCN & 0.295 & 0.207 & 0.146 & 0.184 & 0.138 & 0.076 & 0.122 & 0.106 & 0.076 \\
\hline
\multicolumn{10}{c}{VG-RANDOM} \\
\hline
WL kernel & \underline{0.306} & \underline{0.219} & \underline{0.16} & 0.166 & \underline{0.124} & \underline{0.096} & 0.13 & 0.108 & \underline{0.096} \\
SP kernel & 0.292 & 0.204 & 0.144 & 0.11 & 0.09 & 0.064 & 0.079 & 0.068 & 0.064 \\
RW kernel & 0.0 & 0.0 & 0.0 & 0.0 & 0.0 & 0.0 & 0.007 & 0.0 & 0.0 \\
NH kernel & 0.303 & 0.216 & 0.157 & \underline{0.168} & 0.116 & 0.092 & \underline{0.135} & \underline{0.116} & 0.092 \\
GS kernel & 0.0 & 0.0 & 0.0 & 0.002 & 0.0 & 0.0 & 0.05 & 0.0 & 0.0 \\
\hline
Superv. GAT & 0.35 & 0.268 & 0.213 & 0.382 & 0.294 & 0.176 & 0.294 & 0.245 & 0.176 \\
Superv. GIN & 0.327 & 0.243 & 0.186 & 0.308 & 0.228 & 0.144 & 0.245 & 0.205 & 0.144 \\
Superv. GCN & \textbf{0.369} & \textbf{0.29} & \textbf{0.236} & \textbf{0.424} & \textbf{0.3} & \textbf{0.2} & \textbf{0.295} & \textbf{0.246} & \textbf{0.2} \\
\hline
GAE GAT & 0.294 & 0.205 & 0.146 & 0.132 & 0.104 & 0.07 & 0.092 & 0.085 & 0.07 \\
GAE GIN & 0.297 & 0.209 & 0.15 & 0.114 & 0.096 & 0.07 & 0.08 & 0.076 & 0.07 \\
GAE GCN & 0.291 & 0.203 & 0.143 & 0.114 & 0.1 & 0.066 & 0.079 & 0.078 & 0.066 \\
VGAE GAT & 0.305 & 0.218 & 0.159 & 0.16 & 0.122 & 0.09 & 0.103 & 0.1 & 0.09 \\
VGAE GIN & 0.304 & 0.217 & 0.158 & 0.152 & 0.118 & 0.088 & 0.11 & 0.101 & 0.088 \\
VGAE GCN & 0.303 & 0.216 & 0.157 & \underline{0.156} & 0.122 & 0.088 & 0.101 & 0.089 & 0.088 \\
GFA GAT & 0.292 & 0.203 & 0.143 & 0.142 & 0.108 & 0.076 & 0.107 & 0.094 & 0.076 \\
GFA GIN & 0.298 & 0.211 & 0.151 & 0.142 & 0.11 & 0.082 & 0.11 & 0.093 & 0.082 \\
GFA GCN & 0.297 & 0.209 & 0.149 & 0.14 & 0.112 & 0.082 & 0.106 & 0.095 & 0.082 \\
ARVGA GAT & 0.305 & 0.218 & 0.159 & \underline{0.156} & \underline{0.13} & 0.092 & 0.111 & \underline{0.103} & 0.092 \\
ARVGA GIN & 0.298 & 0.21 & 0.15 & 0.152 & 0.118 & 0.084 & \underline{0.116} & 0.099 & 0.084 \\
ARVGA GCN & \underline{0.306} & \underline{0.22} & \underline{0.161} & 0.154 & \underline{0.13} & \underline{0.094} & 0.113 & 0.097 & \underline{0.094} \\
\hline
\end{tabular}
\vspace*{-0.95\baselineskip}
\end{table}

\begin{table}[h]
\centering
\caption{Average Number of node, edge and total edits, as well as Average Top-1 GED for counterfactual instances retrieved for VG-DENSE and VG-RANDOM.}
\label{tab:combined_metrics}
\vskip 0.1in
\begin{tabular}{c|ccc|c|ccc|c}
\hline
\multirow{2}{*}{Model} & \multicolumn{4}{c|}{VG-DENSE} & \multicolumn{4}{c}{VG-RANDOM} \\
\cmidrule(r){2-5} \cmidrule(r){6-9}
 & Node $\downarrow$ & Edge $\downarrow$ & Total $\downarrow$ & GED $\downarrow$ & Node $\downarrow$ & Edge $\downarrow$ & Total $\downarrow$ & GED $\downarrow$ \\
\midrule
WL kernel & 5.24 & 11.668 & 16.908 & 130.146 & 13.44 & 11.946 & 25.386 & 192.824 \\
SP kernel & \underline{4.964} & 12.078 & 17.042 & \underline{129.334} & 14.144 & 12.386 & 26.53 & 208.096 \\
RW kernel & 9.162 & 20.006 & 29.168 & 217.684 & 17.814 & 22.36 & 40.174 & 385.726 \\
NH kernel & 5.348 & \underline{11.264} & \underline{16.612} & 141.18 & \underline{12.568} & \underline{11.714} & \underline{24.282} & \underline{190.376} \\
GS kernel & 6.444 & 16.748 & 23.192 & 178.434 & 17.216 & 15.344 & 32.56 & 272.744 \\
\hline
Superv. GAT & 5.278 & 10.892 & 16.17 & 108.636 & 12.81 & 11.978 & 24.788 & \textbf{159.324} \\
Superv. GIN & 5.106 & 10.766 & 15.872 & 116.174 & 12.792 & \underline{11.394} & \underline{24.186} & 173.72 \\
Superv. GCN & \underline{4.942} & \textbf{10.312} & \textbf{15.254} & \textbf{105.194} & \underline{12.392} & 11.84 & 24.232 & 159.356 \\
\hline
GAE GAT & 5.504 & 11.848 & 17.352 & 143.976 & \textbf{11.952} & 11.426 & \textbf{23.378} & 197.626 \\
GAE GIN & 5.4 & 12.002 & 17.402 & 143.182 & 12.554 & 11.738 & 24.292 & 201.638 \\
GAE GCN & 5.178 & 11.27 & 16.448 & 143.506 & 12.134 & \textbf{11.384} & 23.518 & 202.638 \\
VGAE GAT & 5.03 & 11.4 & 16.43 & 130.124 & 12.884 & 12.376 & 25.26 & 197.226 \\
VGAE GIN & 4.918 & 10.97 & 15.888 & 131.514 & 12.692 & 12.042 & 24.734 & 199.34 \\
VGAE GCN & 5.114 & 11.702 & 16.816 & 132.056 & 12.91 & 12.392 & 25.302 & \underline{195.938} \\
GFA GAT & 5.07 & 11.564 & 16.634 & 131.582 & 12.988 & 12.412 & 25.4 & 198.78 \\
GFA GIN & \textbf{4.856} & \underline{10.786} & \underline{15.642} & 129.856 & 12.916 & 12.218 & 25.134 & 200.974 \\
GFA GCN & 5.03 & 11.432 & 16.462 & 132.046 & 12.696 & 12.338 & 25.034 & 201.038 \\
ARVGA GAT & 5.062 & 11.482 & 16.544 & 133.476 & 12.508 & 12.192 & 24.7 & 196.946 \\
ARVGA GIN & 4.894 & 10.912 & 15.806 & \underline{128.944} & 12.234 & 11.794 & 24.028 & 198.654 \\
ARVGA GCN & 5.048 & 11.316 & 16.364 & 132.294 & 12.464 & 12.088 & 24.552 & 198.46 \\
\hline
\end{tabular}
\vspace*{-0.95\baselineskip}
\end{table}

In this experimental section, we aim to provide a thorough comparison between different kernel/GAE/GNN architectures as components of the counterfactual retrieval process. We examine them both in terms of their adherence to ground truth rankings, as well as their ability to provide minimal edits regarding both quantity and cost. Throughout this analysis, we inherently contrast supervised and unsupervised GNNs, weighing the trade-offs between employing more efficient versus accurate models and quantifying the resultant incurred damages.

\subsection{Experimental setup}

\paragraph{\textbf{Data}} In all our experiments, we select Visual Genome (VG) \cite{krishna2017visual} as the experimental dataset, due to its abundance in real-life scenes accompanied by manually annotated scene graphs. In addition, VG offers an effortless connection with WordNet, as synset annotations of objects and roles are also provided by annotators. %This connection was implemented via the usage of NLTK \cite{BirdKleinLoper09} python package API\footnote{https://www.nltk.org/howto/wordnet.html}; the same package offers the simple calculation of path similarity scores between two WordNet synsets, providing a number in the [0, 1] range, which denotes the proximity of the synsets within the hierarchy. We repeat the path similarity calculation for all possible node pairs present in selected VG graph pairs $G_{i_r}$, $G_{i_c}$. This way, the replacement (R) cost is defined, by feeding the $1 - path\_similarity$ value to the ground truth GED (Eq. \ref{eq:eq1}). In a similar way, deletion (D) and insertion (I) scores are calculated by finding the shortest path between each $G_{i_r}$ node and the root concept of WordNet (entity.n.01).
Since graph edges are important in our analysis, we regard two density-based dataset splits for training \cite{dimitriou2024structure}: the first one, called VG-DENSE, includes graphs with many interconnections and fewer isolated nodes. The second one called VG-RANDOM does not impose any restrictions on scene graph density. Both splits contain $N$=500 graphs.
As VG does not contain ground truth scene labels we assign them using a pre-trained PLACES-365 classifier \cite{places}. %Counterfactual images may belong to any class defined from PLACES-365, as long as the \textit{query} and the \textit{target} classes differ.

\paragraph{\textbf{Graph Models}} In our study, we employ various graph kernels \cite{kernels-survey} including Shortest Path (SP), Weisfeiler-Lehman (WL), Neighborhood Hash (NH), Random Walk (RW), and Graphlet Sampling (GS), implemented using the GraKel library. For supervised Graph Neural Network (GNN) models, we utilize single-layer siamese GCN, GAT, and GIN as described in \cite{dimitriou2024structure}, trained over 50 epochs with a batch size of 32 and Adam optimizer. Regarding Graph Autoencoders (GAEs), we implement vanilla GAE, Variational GAE (VGAE) \cite{kipf2016variational}, and Adversarially Regularized Variational Graph Autoencoder (ARVGA) \cite{arvga}, alongside a variation of Graph Feature Autoencoder (GFA) \cite{gfa}. GAE and VGAE are both based on the generic auto-encoder structure, but differ in terms of loss function. ARVGA employs basic concepts from Generative Adversarial Networks to further boost the regularization of the latent embeddings produced by VGAE. Finally, GFA utilizes the Feature Decoder (whose goal is to predict the feature matrix $X$), alongside the original VGAE Inner-Product Decoder (whose goal is to predict the adjacency matrix $A$). All these models incorporate single-layer GCN, GAT, and GIN architectures in the encoder (and the decoder for GFA), employing AdamW optimizer with default parameters and a learning rate of 0.001, trained for 20 epochs with a batch size of 32.

GNNs and GAEs were implemented using PyTorch Geometric \cite{pygeo}. In all cases, cosine similarity was used as a measure for embedding similarity. For each scene graph in the dataset acting as the \textit{query}, we retrieve the top $k$ closest graphs with different PLACES-365 labels using a pretrained PLACES classifier with a ResNet50 backbone, resulting in slightly improved labels compared to \cite{dimitriou2024structure}. We then evaluate whether the counterfactual graph, determined by the ground truth GED, is found within this ranked list of $k$ items.

\paragraph{\textbf{Evaluation}}
Information retrieval metrics are used for evaluation, specifically Precision (P@k) and Normalized Discounted Cumulative Gain (NDCG@k). P@k returns the percentage of the relevant items, when considering the first $k$ items in the rank. 
% All the top-$k$ GED-retrieved items are considered to be relevant and equally weighted. 
We also employ the ``binary'' variant from \cite{dimitriou2024structure}, where only the top-1 GED item is relevant, in line with the counterfactual logic. Binary NDCG@k compares the rank of the top-1 counterfactual returned by an algorithm to the ideal ground truth rank obtained by GED.
% as in Eq. \ref{eq:eq1}
While NDCG@k lacks direct interpretability, it's widely used for comparing retrieval system performance. Our evaluation focuses on $k$=1, 2, 4 ranks, once again emphasizing proximity to the top-1 result. Additional metrics assess the proposed edits' minimality by GNNs, including average node, edge, and overall edits, alongside average GED of counterfactuals. Lower values indicate better model performance, with these metrics considered relative to each other.

\subsection{Quantitative Results}
In Table \ref{tab:combined-results}, we present ranking results regarding counterfactual retrieval using all proposed graph models: graph kernels, supervised GNNs and unsupervised GAEs. For both dataset subsets, all supervised models were trained using 70K scene graph pairs extracted from $N$=500 graphs, while all unsupervised models were trained on the same 500 graphs (without pairing them).

Kernels reveal distinct performance patterns, with some closely rivaling unsupervised autoencoders. Notably, in VG-DENSE, SP and WL kernels excel, while in VG-RANDOM, WL and NH show superior performance. However, SP and NH metrics falter in VG-RANDOM and VG-DENSE, respectively, underscoring the significance of edge density. Conversely, kernels like RW consistently fail to retrieve counterfactual graphs, indicating their limitations in capturing graph structure. Supervised GNN models, particularly GCN, outperform kernels in ranking metrics, benefiting from optimized learning on ground truth GED. Unsupervised autoencoders exhibit more variability, with ARVGA GIN excelling in VG-DENSE and VG-RANDOM splits, underlining the importance of adversarial regularization for competitive performance.
% Regarding the non-monotonic behavior of NDCG, we observe that the score does not necessarily decrease as we include more retrieved objects. This phenomenon is commonplace in the context of Information Retrieval tasks, and it reflects the system’s capability to compensate for potential inaccuracies at the top rank by retrieving a well-ordered and relevant set of objects within a broader scope. 
% For example, a model could score better on NDCG@4 than NDCG@1, in the case where the top-retrieved object is incorrect, yet the subsequent objects in the ranking prove to be highly relevant to the ground truth rank.

In Table \ref{tab:combined_metrics}, we present results on the average number of edits, a common metric for evaluating counterfactual explanation systems, alongside the average top-1 counterfactual GED. In VG-DENSE experiments, the superior performance of supervised GCN is evident in edit numbers. While total edits appear similar across models, GFA GIN emerges as the second-best performer, indicating the limitations of relying solely on edit counts without considering their cost (GED). Comparing supervised GCN and GFA GIN, a significant gap in GED underscores their semantic divergence. Results for VG-RANDOM exhibit more variability, with GAE GAT leading in total edits. Considering both edit counts and GED provides insights into explanation minimality, highlighting the semantic relevance advantage of supervised models despite similar edit counts.

By comparing ranking metrics across the two dataset splits, we can easily observe that best results overall were achieved in the DENSE split. This is an expected behavior, since GNNs require interconnections for message passing procedures, i.e. to exchange and aggregate information from their neighbors. 

\paragraph{\textbf{Performance vs dataset size}}
Although unsupervised GAEs lag behind supervised models, there is a notable discrepancy in the required ground truth data. Supervised GNNs necessitate training on $\sim N^2/2$=70K pairs, maintaining a quadratic relationship with input data, entailing computationally expensive ground truth GED calculations. Conversely, unsupervised GAEs offer a more scalable solution, requiring $N$ training samples without optimizing on ground truth GED. This enables faster calculation of counterfactual explanations. While increasing training samples can enhance GAE performance, it may not be feasible in low-resource scenarios or where manual graph construction is required. Both approaches demonstrate computational efficiency, with GAE training taking less than 2 minutes for $N$=500 graphs, whereas supervised GNN training requires $\sim$3 hours for $\sim N^2/2$=70K graphs, with slight variations based on graph density. Despite differing training times, the retrieval, inference, and edit path computation times remain consistent between the two approaches. Training on a single GPU available in online platforms e.g. Kaggle or Google Colab is feasible for both supervised and unsupervised GNNs, while retrieval and inference operations are CPU-based, ensuring reproducibility of reported results.

\subsection{Qualitative Results}

\begin{figure*}[h!]
    \centering
    \includegraphics[width=0.95\textwidth, height=1.4\textwidth]{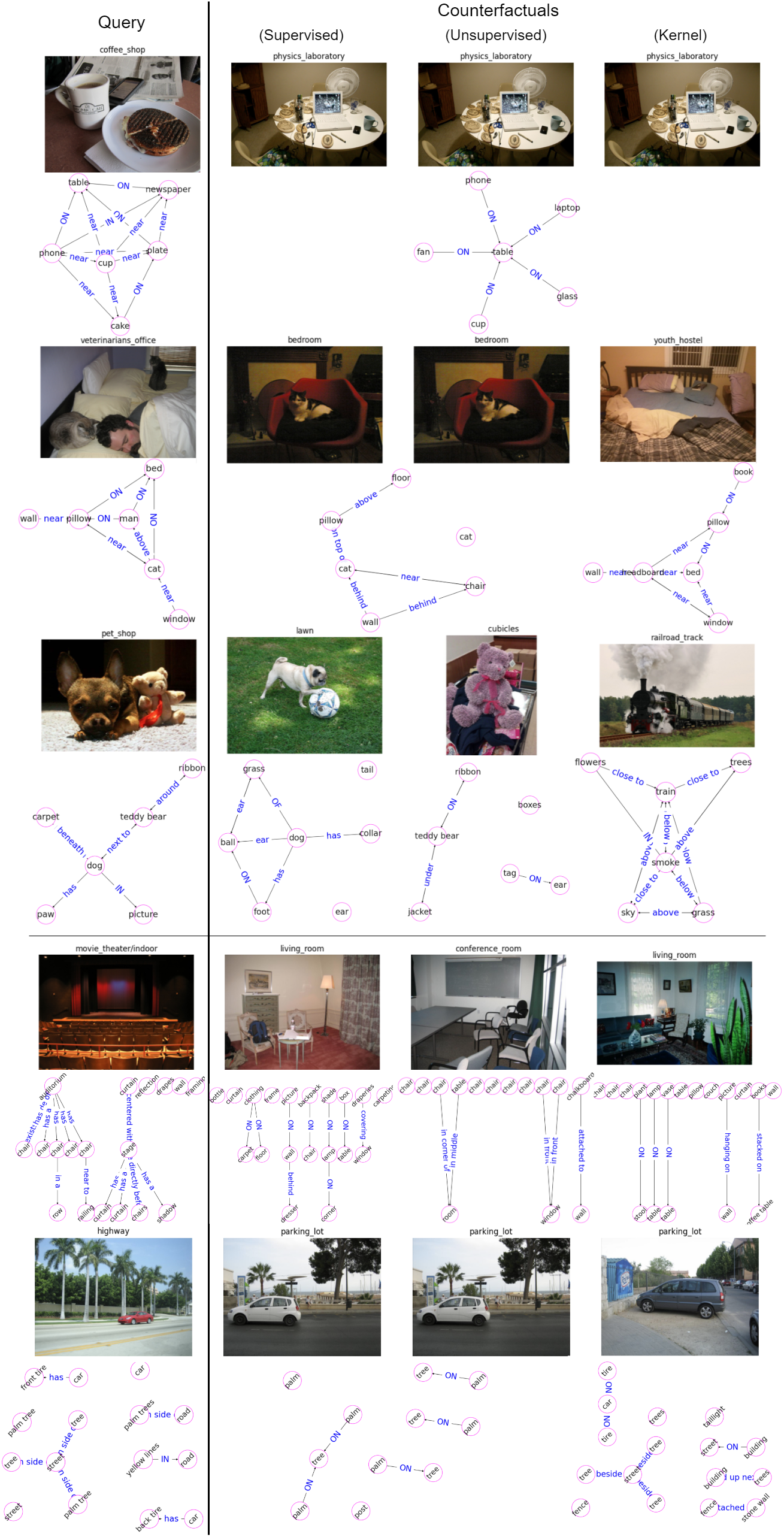}
    \caption{Counterfactuals from the best supervised GNN, unsupervised GNN and kernel for VG-DENSE (top 3) and VG-RANDOM (bottom 2).}
    \label{fig:qualitative}
\vspace*{-0.9\baselineskip}
\end{figure*}

Qualitative analysis is indispensable, particularly in the visual domain, where numerical metrics may not fully capture model performance. For graph retrieval, structural and semantic cues play vital roles. In Figure \ref{fig:qualitative}, we present results from top-performing models: supervised GCN, unsupervised ARVGA-GIN, and WL-kernel. Note that counterfactual classes do not need to align between models.

% A qualitative analysis is crucial, as our counterfactual application lies in the visual domain. Even though some models may present similar ranking metrics, we cannot conclude about their quality according to visual perception from merely observing numerical values. This fact is essential in the case of graphs, as either structural or semantic-based cues may be more influential.

% \vspace*{-0.99\baselineskip}

% In Figure \ref{fig:qualitative}, we showcase qualitative counterfactual results employing kernels, as well as both supervised and unsupervised GNNs. The instances presented were acquired from the top-performing models in each scenario, specifically the supervised GCN, unsupervised ARVGA-GIN, and WL-kernel. Note that counterfactual classes do not need to align between models.

In the case of VG-DENSE, multiple instances exhibit concordance among models, owing to its dense interconnections and the inherent ability of all models to recognize matching structures. The 1st row demonstrates unanimous agreement, as the obtained scene graph essentially forms a subgraph of the query. A human could also agree on the semantic closeness between counterfactual instances: 
% The query image corresponds to a "kitchen" table, where a person can have breakfast, while drinking coffee and reading their newspaper. The retrieved instances denote a "workstation" table, where a person can work on their laptop while also drinking a cup of coffee. 
Semantics such as ``table'', ``cup'', ``phone'' are preserved, while the kitchen-related ``cake'' and ``newspaper'' should be edited to allow workstation-related semantics, such as ``laptop''. Moving to the 2nd row, GNN methods align; successfully integrating structural and semantic information. Despite the kernel-retrieved result having a more similar structure, GNN methods successfully discern and retrieve an image featuring a cat on furniture, surpassing a mere depiction of a bed.
% : the knowledge-based semantics allow the information of the WordNet hierarchy \textit{bed-isA-furniture} and \textit{chair-isA-furniture} to drive counterfactual retrieval.
In terms of human perception, counterfactuals depicting a ``bedroom'' (query) and a ``living room'' instance (GNN-based retrievals) respect the change of class label (``bedroom''$\rightarrow$``living room'') with minimally altering semantics (``pillow'' and ``cat'' semantics are preserved), allowing effortless high-level interpretability. Conversely, in the 3rd row with a widespread disagreement, the merits of the supervised GNN network become evident. The unsupervised GAE captures an image resembling the query with a teddy bear wearing a bow but overlooks the presence of a dog.
% , leading to decreased preservation of semantics.
In contrast, the supervised GNN, having more accurately approximated GED, retrieves an image centered on a dog and a different type of toy.
% , while altering the class from "living room" to "garden". 
% Although the toy may differ, the semantic proximity of the pug image is notable. 
% Once again, hierarchical knowledge allows this abstraction (\textit{teddy bear-isA-toy} and \textit{ball-isA-toy}, offering high-level interpretability of suggested counterfactuals. 
Interestingly, the kernel fails and retrieves an image of a train, emphasizing that a sole focus on structure  (star subgraph) is insufficient. 
% While the kernel aligns with the star shape of the query's scene graph, the interconnected objects differ entirely, and the depicted scenes are totally unrelated.

Assessing retrieved VG-RANDOM results proves challenging due to the dissimilar nature of the sparser underlying graphs. In the 4th row, the supervised GCN retrieves a structurally closer scene graph, despite visual similarities in GNN-based counterfactuals. It becomes harder even for humans to evaluate their proximity, since all instances contain ``chair'' and ``curtain'' concepts, but no other discriminative structure or detail. In the final row the models reach consensus, successfully retrieving an image with palm trees, a characteristic the kernel fails to achieve despite the retrieval of very closely aligned graphs. 
% The closeness of VG-RANDOM results highlights that a GNN counterfactual framework requires rich interconnections to perform in a meaningful manner. At the same time, interconnections are indeed meaningful to humans, as proven by VG-DENSE examples, where evaluation of counterfactuals was easier and more interpretable, in contrast to VG-RANDOM retrievals. This observation aligns with the question regarding the relatedness between expressive scene graphs and human intuition, since concept interconnections may be highly informative (as in the case of the running example of Figure \ref{fig:example-graph}). To our favor, data preprocessing can eliminate sparser graphs from our analysis, enabling more informative explanations.
% The effectiveness of a GNN counterfactual framework, as observed in VG-RANDOM results, relies heavily on rich interconnections. VG-DENSE examples underscore the necessity of meaningful interconnections for both model performance and human interpretability, in contrast to VG-RANDOM retrievals. This emphasizes the significance of expressive scene graphs and human intuition, where the interplay of conceptual connections provides valuable insights, as shown in Figure \ref{fig:example-graph}. Furthermore, thoughtful data preprocessing holds potential for eliminating sparser graphs, thereby enhancing the informativeness of explanations.
The effectiveness of the framework relies on rich interconnections observed in VG-DENSE, underscoring the significance of expressive scene graphs, human intuition, and thoughtful data preprocessing for enhancing explanation informativeness.

Notably, none of the retrieved instances are mislead by purely visual characteristics, such as color, contrast or brightness of images. This is because graph-based algorithms naturally disregard any such characteristics, as they are not integrated into scene graphs.
% , allowing a more human-related interpretation. 
% This fact becomes evident if we think about e.g. a kitchen instance, such as the one in the first row. Even if we convert this image to grayscale or if we increase/decrease brightness, it still remains an image of a kitchen, depicting the same objects. 
% A human could effortlessly insist that the new image still depicts the same image under all these conversions, concluding that graph machine learning models, and especially GNNs are capable to fully align with human perception.
Using the kitchen scenario as an example, the preservation of its identity through alterations like converting to grayscale or adjusting brightness suggests that graph machine learning models, particularly GNNs, can closely align with human perception.

\section{Conclusion}
In this work, we unify graph-based conceptual counterfactual explanation algorithms under a common framework, upon which we conduct a comparative study.
% A variety of graph based algorithms, including graph kernels, supervised and unsupervised graph neural networks demonstrate the merits of graph-based representations for counterfactuals, providing insightful information on what needs to be changed conceptually to transit to another classification label in a computationally efficient manner, while offering optimality guarantees from hierarchies.
Various graph-based algorithms, such as kernels, supervised and unsupervised GNNs, highlight the value of graph representations for counterfactuals, offering insights on what needs to be changed conceptually to transit to another class efficiently.
%, which significantly accelerates the graph matching procedure without exhaustively calculating graph edit distance for all graph pairs of the dataset. 
Quantitative and qualitative analysis showcases the different criteria that influence the retrieval of counterfactual instances from varying graph algorithms, and the importance of well-defined and informative semantics towards human-interpretable explanations.
There is ample room for further real-world applications of this framework within various domains where we have access to graph-structured data, such as recommendation systems, transportation networks, as well as agricultural and health inspection sectors.

%In this work, we introduce a novel framework that employs Graph Machine Learning algorithms that optimize the search of conceptual edits to provide counterfactual explanations of images. Scene graphs are leveraged as a means for conceptual representation of a given image, highlighting the importance of interconnections, which was missing from previous conceptual counterfactual frameworks. 
%Under this abstraction, searching for counterfactual scenes reduces to a graph matching problem, while knowledge-based edit costs suggest the conceptually minimum changes needed to be made on a given graph to change its classification label.
%A variety of graph based algorithms, including graph kernels, supervised and unsupervised graph neural networks demonstrate the merits of our proposed framework, which significantly accelerates the graph matching procedure without exhaustively calculating graph edit distance for all graph pairs of the dataset. Quantitative and qualitative analysis showcases the different criteria that influence the retrieval of counterfactual instances from varying graph algorithms, and the importance of well-defined and informative semantics towards human-interpretable explanations. Overall, our conceptual counterfactual framework provides insightful information on what needs to be changed conceptually to transit to another classification label, providing optimality guarantees from knowledge bases and computational efficiency from utilizing lightweight graph models.

\begin{credits}
\subsubsection{\ackname}
This research work is Co-funded from the European Union’s Horizon Europe Research and Innovation programme under Grant Agreement No 101119714 — dAIry 4.0.

\subsubsection{\discintname}
The authors have no competing interests to declare that are
relevant to the content of this article.

\end{credits}


\bibliographystyle{splncs04}
\bibliography{main}





% LNCS Homepage, \url{http://www.springer.com/lncs}. Last accessed 4
% \end{thebibliography}

\end{document}